\documentclass[preprint,12pt,3p, export]{elsarticle}

\usepackage{amssymb, listings, adjustbox, longtable, tabu}

\journal{6.863 Final Project}

\begin{document}

\begin{frontmatter}

\title{Natural Language and Spatial Rules \tnoteref{label0}}
\tnotetext[label0]{Massachusetts Institute of Technology, Department of Electrical Engineering and Computer Science, 6.863J/9.611J Natural Language Processing}

\author[label1]{Alexandros Haridis}
\address[label1]{PhD Student, Department of Architecture, MIT}
\ead{charidis@mit.edu}

\author[label2]{Styliani Rossikopoulou Pappa}
\address[label2]{Graduate Student, Master in Design Studies, Harvard Graduate School of Design }
\ead{spappa@gsd.harvard.edu}

%\author[label3]{Emily Stanford}
%\address[label3]{Undergraduate Student, Department of %Electrical Engineering and Computer Science}
%\ead{emistan4@mit.edu}

\begin{abstract}
We develop a system that formally represents \emph{spatial semantics} concepts within natural language descriptions of shapes given in text. The system builds upon a model of spatial semantics representation according to which words in a sentence are assigned spatial roles and the relations among these roles are represented with spatial relations. We combine our system with the shape grammar formalism that uses \emph{shape rules} to generate arrangements of shapes. Our proposed system then consists of pairs of shape rules and \emph{verbal rules} where verbal rules describe in natural language the \emph{action} of a corresponding shape rule. We present various types of natural language descriptions of shapes that are successfully parsed by our system and we discuss open questions and challenges.
\end{abstract}


\end{frontmatter}

% \linenumbers

\section{Introduction}
\label{sec1}

\noindent Natural language is generally recognized as one of the important human faculties capable of expressing spatial information and visual material ~\cite{LandauJacken, WinstonDirPerc, Jackendoff2012, TenbrinkEds, Ungerleider, Landau, MBhatt}. Language provides verbal tools for the schematization of space and for describing the visual appearance or structural characteristics of perceived objects ~\cite{TenbrinkEds, TalmyVol1, Winograd, MBhatt}. A number of questions arise in this regard. Can natural language substitute for perceptual information? What are some of the situations where this may indeed happen? What are some of the cases where such a task is impossible? In this project, we propose to investigate this interface between natural language and perception of visual material and to evaluate whether, and to what extent, sensory and linguistic stimuli support the formation of equivalent spatial representations.\par
This present write up is organized in the following way. Section 1 continues with an overview of selected literature on research conducted from a variety of knowledge areas and disciplines on the interface between natural language and perception of visual-spatial material. In Section 1.2, we describe our methodology by briefly explaining the basic elements of our computational approach and we outline the intended contributions of our project. In Section 2, we describe a model of spatial semantics representation that we have used to build our system. In section 3, we briefly describe shape grammars and how shape rules are defined and used to generate languages of shapes. In section 4, we give a motivating example that illustrates how shape rules are combined with verbal rules to build natural language descriptions of arrangements of shapes and their corresponding spatial semantic representation. In section 5, we outline the formal spatial semantics representation based on the lambda calculus method and show various examples that illustrate the capabilities of our system. In section 6, we discuss open questions, challenges and interesting observations we made throughout the development of this project. Finally, we provide an Appendix section with supporting material.

\subsection{Background}
\label{subsec1}

\noindent The study of verbal descriptions for spatial material, such as shapes, diagrams, pictures, drawings, physical objects, physical spaces, and so on, can be divided into two broad directions: verbal descriptions that attempt to capture \emph{space} (that is, physical space, an environment, a natural scene, and so on), and verbal descriptions that attempt to capture \emph{object} (that is, a physical-material object, an entity in a scene, objects in an environment, shapes in pictures, and so on). While in our project we exclusively focus on the second type of verbal descriptions, in this background section we cover both types because we judge that it is important to have a good sense of what kind of work has been conducted at the interface between natural language descriptions and perception of visual-spatial material.\par
In the literature of experimental psychology, we very commonly find studies that examine linguistic representations of physical environments. A classic example is the 1975 work of~\cite{LindeLabov} who asked New Yorkers to answer the question, "Could you tell me the lay-out of your apartment?" The majority of the participants in their study gave verbal descriptions based on imaginary tours through their apartment. In general, we may distinguish three common ways of experiencing a certain environment with the purpose of constructing a verbal description of it: walking through the environment or \emph{route description}; standing in one place in the environment or \emph{gaze tour description}; viewing the environment 'from above' or \emph{survey perspective}.
This classification is given in~\cite{TenbrinkEds}. An important point worth raising is that physical environments (like an apartment) are multi-dimensional. Not only because they are entities in a three-dimensional world but also because they can be characterized in terms of a variety of things, such as geometry, color, materials, light, etc. However, no matter the environment picked, the linguistic representation of the environment has to be inherently one dimensional ~\cite{TenbrinkEds, TverskyEtAl94, HabelTappe}. That is to say, if someone wants to construct a verbal description of a physical environment (and indeed of any physical object in general) then one needs to \emph{linearize} space into language or in other words one needs to lexicalize space in the terms of a one-dimensional representational system, namely, natural language. This important aspect of natural language holds a central place in our project where we attempt to build natural language descriptions of the appearance of shapes, of how a given shape or arrangement of shapes 'looks'. In this respect, some of the questions that arise are, which aspects of an object come first in a linearized description of the object? Which aspects are emphasized at the expense of ignoring others? Why is one description better than another? Can there even be an adequate description of the object?\par
The connections between formal models of computation of space, and 'space' as it occurs in language are themselves a matter of intense research activity. This is particularly evident in the computer science disciplines concerned with the investigation of an artificial intelligence capable of 'talking' about space or about perceived objects within space. For example,~\cite{MBhatt} develop an architectural design assistant with a broader focus to examine the relationship between the conception, formalization, and the computational aspects of space as it occurs within systems of human assistance. Such systems are meant to know the properties of physical space and are skilled in dealing with the spaces they 'know' in such a way so that they support humans, such as interior space designers, engineers, media designers, and so on~\cite{MBhatt}. The relationship between language and perception of visual-spatial material was of great interest already in the 1970s when the first artificial intelligence programs started to emerge. For example, Winograd's system~\cite{Winograd} was a system that could understand verbal instructions in English about specific spatial tasks within a 'blocks-world' setting, such as "pick up a big red block" or "what is the pyramid supported by?". Winston's system ~\cite{WinstonPhD} was building and understanding descriptions of physical objects such as an "arch" or a "house" and it used a form of spatial language where relations between objects where modeled using notions, such as left-of, on-top-of, part-of, behind. Another early work was the system of Boberg ~\cite{BobergMS}, which was an early attempt to build a machine 'envisioner' that could understand verbal descriptions of spatial objects and draw the described objects. Like Winograd's system, Boberg's system was also situated within a 'block's world'. In our project, we focus not on a block's world where objects have predefined geometric and/or material characteristics but we focus on shapes and arrangements of shapes whose visual appearance is ambiguous and no predefined set of component parts exists. This interface between perception and natural language still holds an important role (and largely open) in artificial intelligence. This can be seen from many recent attempts to create artificial intelligence systems that can describe images as humans do. This task is mainly approached from an engineering standpoint and typical tasks involved in research are image captioning and annotation, question answering, and recognition and description of objects and relationships among them. For example, two recent efforts towards these ends are the Visual Genome dataset in~\cite{krishnavisualgenome}, a large dataset that aims to connect language and vision by gathering dense image annotations via crowdsourcing, and CLEVR~\cite{Clevr}, a dataset that is aimed at supporting systems in elementary visual reasoning tasks with an emphasis on compositional reasoning (i.e. the ability to understand how a complex scene is composed out of simpler parts). \par   
The importance of language in describing visual-spatial material has been recognized in design research in architecture and the visual arts. For example, the now classic (self)study of artist Paul Klee~\cite{Klee} shows very clearly that verbal descriptions are important for pedagogical purposes, in particular, in how to teach someone to talk about the visual, spatial characteristics of a painting or of any type of aesthetic object. Language-enabled descriptions are pervasive in architectural design education and in particular in the studio culture as the analytical study of ~\cite{Schon} shows. In the more formal, computational approaches towards design, ~\cite{StinyShapeBook} and ~\cite{StinyDescriptions} have shown how the shape of aesthetic objects, such as buildings, paintings, designs, can be computed with a shape grammar and how the spatial characteristics of these objects can be described using special forms of description grammars. In our project we use the shape grammar formalism because it enables us to describe arrangements of shapes in a visual manner through shape rules (see section 2). \par
Other relevant work to our project, is the research conducted on spatial semantics~\cite{Zlatev, Kordjamshidi1, Kordjamshidi2} where the goal is to understand spatial expressions in natural language. Two of the most common notions in terms of which spatial language is formalized is topology and orientation. There exist special forms of calculus systems that compute topological relations between geometric objects called regions~\cite{ManiPustejovsky} and these relations are represented by prepositions in language. Work conducted towards this direction aims at developing systems that can infer spatial relations between either abstract or real objects in scenes by reading a natural language description in text. The systems then are able to annotate the given sentences by inferring spatial prepositions, actors, trajectors, and other thematic roles. Some recent work conducted in this direction from which we take inspiration in this project are ~\cite{Kordjamshidi1, Kordjamshidi2, SpatialML}. Special forms of natural language have been also developed in order to describe more specific spatial phenomena, such as for example how language may capture the way spatial entities interact with respect to force~\cite{TalmyVol1}, or how language may be used to describe movement~\cite{ManiPustejovsky}.\par

\subsection{Project Summary and Intended Contributions}
\label{subsec1}

\noindent In this project, we develop i) a \emph{model of spatial semantics representation} that interprets a sentence given in text according to basic spatial semantics concepts; ii) a \emph{methodology} for combining shape rules and verbal rules for the generation of specific languages of shapes as well as the generation of corresponding verbal descriptions; and iii) an \emph{implementation} of our model based on a rule-to-rule association of context-free rules and lambda calculus forms. With this project, our goal is to highlight some of the limitations of language to capture visual-spatial phenomena that a human 'eye' would capture with no particular difficulty. To show some of the challenges that one may face when approaching the task of capturing spatial semantics concepts from a given natural language description in text. Finally, to demonstrate the elements of a system that successfully captures spatial roles and spatial relations among them and associates them with a visual arrangement of shapes.

\section{Spatial Semantics Concepts}
\label{sec1}

\subsection{Verbal description styles: constructive and 'from-above'}
\label{sec1}

\noindent Given an arrangement of shapes, there are at least the following two verbal description styles that one may follow to construct a natural language description of the arrangement: \emph{constructive} and \emph{'from above'}. A constructive description tells how the arrangement can be constructed, i.e. drawn on a paper, in a step by step manner. A constructive description is different from a $‘perspective’$ or $‘from-above’$ description according to which one describes an arrangement in a static way, emphasizing certain perceived parts. From-above descriptions have a declarative nature; they are meant to tell how things are. Constructive descriptions tell how things can be derived in a step by step manner. \par
In our project, we only consider specific classes of arrangements of shapes (see sections 3 and 4). For each arrangement of shapes, we have built an associated grammar that consists of shape rules and corresponding verbal rules. These grammars provide, almost naturally, a framework for generating constructive descriptions of arrangements of shapes. However, the two description styles, namely, constructive and from-above, can be easily converted to one another. In particular, constructive descriptions can be characterized as verb phrases that always start with an action verb, such as 'add', 'put', or 'draw'. From-above descriptions make use of 'is' to declare a fact or assertion. Hence, one may convert a from-above description to a constructive description by simply adding an action verb and removing all existential verbs like 'is'. To give an example, consider the from-above description, \par

\vspace{0.2in}

\emph{“the square is next to the rectangle.”}\par 

\vspace{0.2in}

\noindent This sentence can be converted into a constructive description in a straightforward way, like so, \par 

\vspace{0.2in}

\emph{“add the square next to the rectangle.”} \par

\vspace{0.2in}

\noindent We could likewise use the verbs 'draw' or 'put' (or similar ones). In our formal spatial semantics representation (section 5), we handle both description styles although their underlying structures (and in particular, their event structures) do not differ much as expected.\par

\subsection{Basic spatial semantics concepts}
\label{sec1}

\noindent Spatial semantics concepts are intermediaries between perception and natural language. They provide cross-linguistic spatial concepts for specifying how entities in (two or three dimensional) space are arranged. We use spatial semantics concepts mainly as a general framework for organizing the spatial roles that the words in a verbal description sentence play and how these spatial roles are related to one another within the sentence. The content of this subsection draws from a variety of sources, in particular~\cite{LandauJacken, Zlatev, Kordjamshidi1, Kordjamshidi2, ManiPustejovsky, RandellRCC8, Levinson}. In our system, we use the following basic spatial semantics concepts:\par

\vspace{0.2in}

\noindent \emph{Trajector}: an entity whose location or motion is of relevance in the description. It can be an object, like a shape, a geometric attribute of the shape, such as an edge or a corner, or it can be itself an event. In psychology, the trajector is sometimes called the 'figure' or the 'referent'.\par

\vspace{0.1in}

\noindent \emph{Landmark}: a reference entity in relation to which the location or the movement of the trajector is specified. In psychology, the landmark is sometimes called the 'ground' or the 'relatum'. \par

\vspace{0.1in}

\noindent \emph{Frame of reference}: a linguistic concept that specifies certain reference points or even a coordinate system based on axes and angles. An object can be a frame of reference, such as the landmark. In English, there are three lexicalized frames of reference: \emph{relative}, \emph{absolute} and \emph{intrinsic}~\cite{Levinson}. An absolute frame of reference, for example, can be the geolocation of an object with respect to the north or south pole. In our project, we are interested in describing spatial arrangements and more specifically to specify a shape with respect to another shape or with respect to an attribute of a shape. Thus, we exclusively use a relative frames of reference. In our case, the frame of reference in each description sentence is implied by the spatial roles of the trajector and the landmark along with a directional system (see below).\par

\vspace{0.1in}

\noindent \emph{Direction}: specifies directional relations with respect to a frame of reference. Frames of reference in our description sentences are relative and are always implied by the spatial roles of the trajector and the landmark. We only use the following relative directions \{ \emph{left}, \emph{right}, \emph{top},\emph{bottom} \}. It is of course assumed that by, for example, 'left edge' we mean the edge of the shape that is to the left with respect to a viewer that looks at the shape drawn on the two dimensional plane.\par

\vspace{0.1in}

\noindent \emph{Spatial locator}: a non-projective, locative preposition that describes the spatial relation between the trajector and the landmark. It takes the following values \{ 'at,' 'in,' 'on' \}. The meaning of a spatial locator is also represented in more detail by a topological relation, which we call 'region' (see below). \par

\vspace{0.1in}

\noindent \emph{Region}: a concept that denotes a topological relation that defines how a trajector, or more generally some region of space, is related to a landmark. The values of a region may denote concepts, such as contiguity, parthood, inclusion, overlap, and others. These values are denoted more elaborately using the rules of the RCC8 regional calculus~\cite{RandellRCC8}. We use the following three types of relations from RCC8:\par

\begin{figure}[h!]
\centering
\includegraphics[scale=1.0]{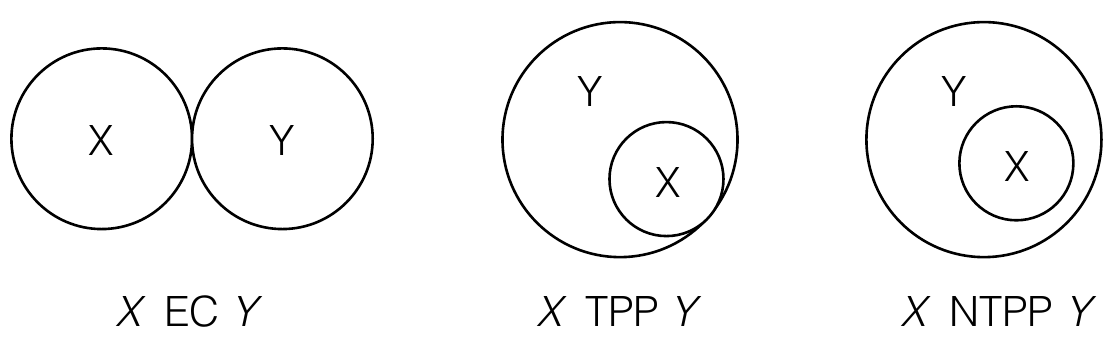}
\label{Figure1}
\end{figure}

\noindent Each regional relation above has a logical specification. More specifically, according to~\cite{RandellRCC8}, a predicate \emph{Connect}(x, y) evaluates to true if the objects x and y are connected in some way, in our case, if shapes x and y 'touch'; it evaluates to false when x and y are not connected in some way, i.e. shapes x and y do not touch. Parthood relation is expressed in the following way. \emph{Part}(x, y) evaluates to true if and only if for every z, \emph{Connect}(z, x) implies \emph{Connect}(z, y). Now we can define overlap. \emph{Overlap}(x, y) evaluates to true if there exists a z, such that \emph{Part}(z, x) evaluates to true and \emph{Part}(z, y) evaluates also to true.\par The relation of overlap can be easily expressed with shape arithmetic: two shapes overlap if they share some part. As a simple illustration, consider the following Figure: \par

\begin{figure}[h!]
\centering
\includegraphics[scale=0.7]{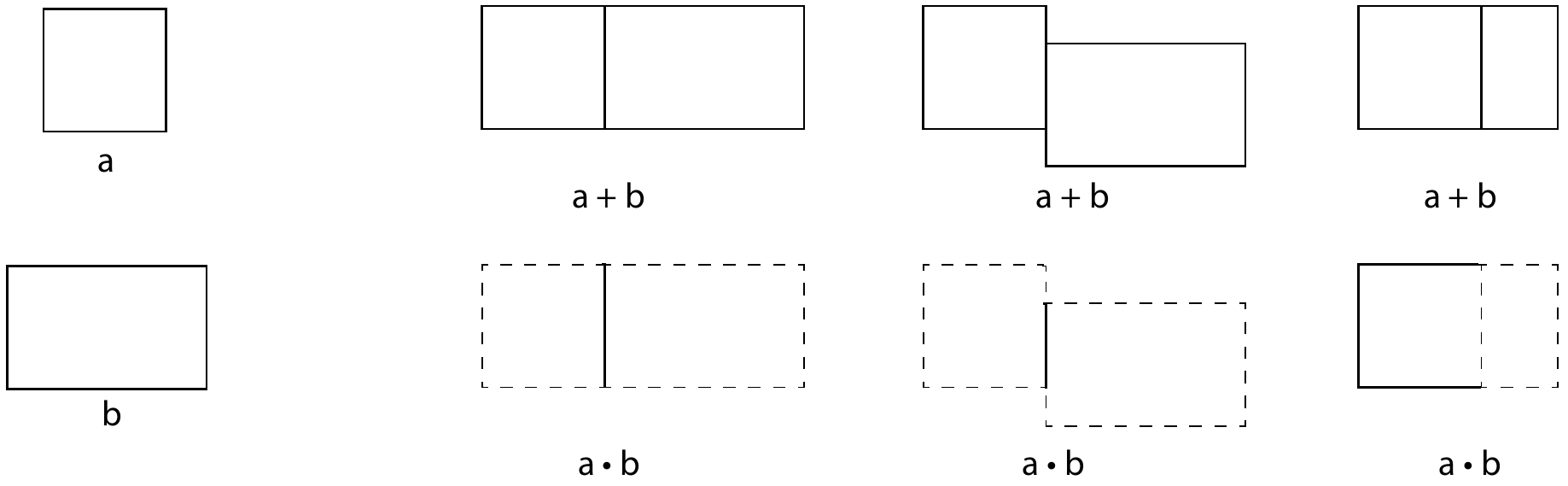}
\label{Figure2}
\end{figure}

\noindent The Figure shows two shapes, labelled \emph{a} and \emph{b} (mainly for ease of reference). In general, two shapes may be put together in many different ways (in fact, infinitely many). In the Figure we only show two such ways. With shapes, the result of adding one shape onto the other can be represented with the arithmetic operation of sum, that is, \emph{a} + \emph{b}. The Figure thus shows three ways of adding shape \emph{b} to \emph{a} (or the converse), such that the two shapes overlap. Now, the overlapping part is simply the multiplication of the two shapes, which computes the common part. Therefore, for all the above three cases, the predicate \emph{Overlap}(a, b) would evaluate to true.\par
Using connect, parthood and overlap, we may define new regional relations. In particular, we define the three relations EC, TPP, and NTPP (as shown in the previous Figure). EC(x, y) means that x is externally connected to y, if Connect(x, y) evaluates to true and Overlap(x, y) is not true. TPP(x, y) means that x is a tangential propert part of y, and NTPP(x, y) means that x is non-tangential propert part of y (we choose to omit their logical forms because they can be derived in a straightforward way, but the interested reader may refer to ~\cite{RandellRCC8}).
\noindent English spatial prepositions 'at', 'in', 'on' can be mapped to regional relations. To do so, we follow ~\cite{ManiPustejovsky}. The preposition 'at', when used spatially denotes a trajector object that coincides in some way with a landmark object, or that it tangentially overlaps some part of the landmark object. Thus, 'at' is mapped to \{TPP, NTPP\}. Using similar reasoning, the preposition 'on' has a meaning of contiguity as well as it may function as a kind of support. For example, "the book is on the table". Thus, 'on' is mapped to \{EC, TPP\}. Similarly, the preposition 'in' typically means that a trajector object is contained in a landmark object. Or it may denote enclosure. Thus, 'in' is mapped to \{EC, TPP, NTPP\}. We should note at this point that this mapping of spatial prepositions to regional relations is quite ambiguous (see ~\cite{ManiPustejovsky} for relevant discussion). Nevertheless, we find it useful to include this information in our system for purposes of completeness but also because it may have some further uses that we shortly outline in the Discussion section.\par

\vspace{0.2in}

\noindent Apart from the above spatial semantics concepts, we introduce two additional concepts that are specific to our project. When describing an arrangement of shapes, it is common to refer to attributes of shapes related to their geometry. For example, the sentence "the right edge of shape1" makes explicit reference to the attribute 'edge' of shape1, and more specifically, among the edges of shape1, it makes reference to the 'right' edge. Other examples of shape attributes are 'midpoint' and 'corner'. To handle sentences that make explicit reference to shape attributes, we introduce the following semantic concepts:\par

\vspace{0.2in}

\noindent \emph{Property of shape}: this acts as an operator that applies to a shape object and extracts geometric information about the shape. In our project, we use the following properties {‘edge’, ‘midpoint’, ‘corner’}. Properties can be combined with information about directionality to distinguish, for example, between the four edges of a square.\par

\vspace{0.1in}

\noindent \emph{Shape}: a direct reference link to a shape entity in the given arrangement of shapes. This is not a string or a structured representation, but the actual ‘picture’ of the referenced shape. We denote referenced shapes in description sentences with angle brackets $< ... >$. When our grammar generates a sentence, the references to specific shapes from the arrangement are replaced with the actual pictures of the referent shapes (see section 4).\par

\section{Shape rules and verbal rules}
\label{sec1}

\subsection{Shapes and shape rules}
\label{sec1}

\noindent A specific class of formal machines useful for generating arrangements of shapes are Shape Grammars (SG)~\cite{StinyShapeBook}. A shape grammar is defined with a set of shape rules. A shape rule has the general form \emph{a} $\rightarrow$ \emph{b}, where \emph{a} and \emph{b} are not words (or symbols) but shapes. Shape rules generate arrangements of shapes that constitute a language. We give a few examples of languages of shapes for purposes of illustration. Consider the following pair of shape rules,

\begin{figure}[h!]
\centering
\includegraphics[scale=0.5]{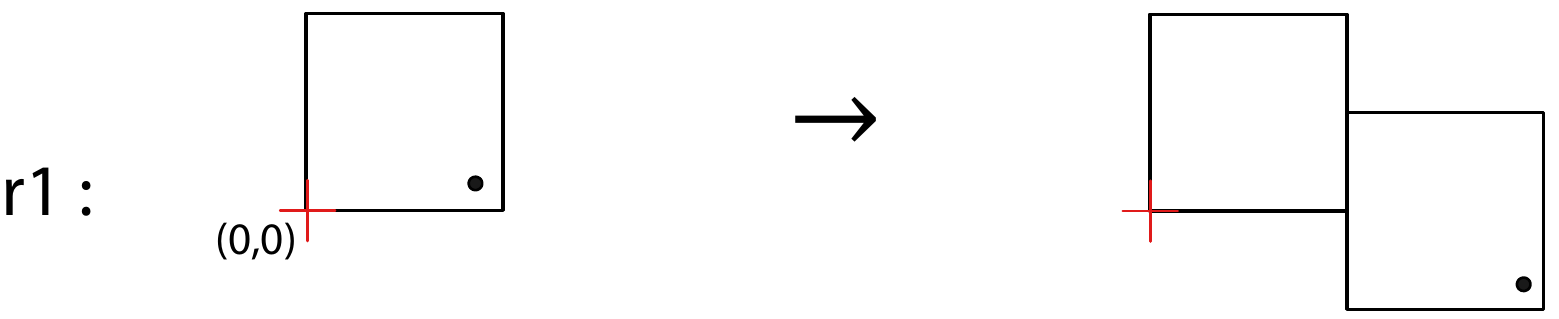}
\includegraphics[scale=0.5]{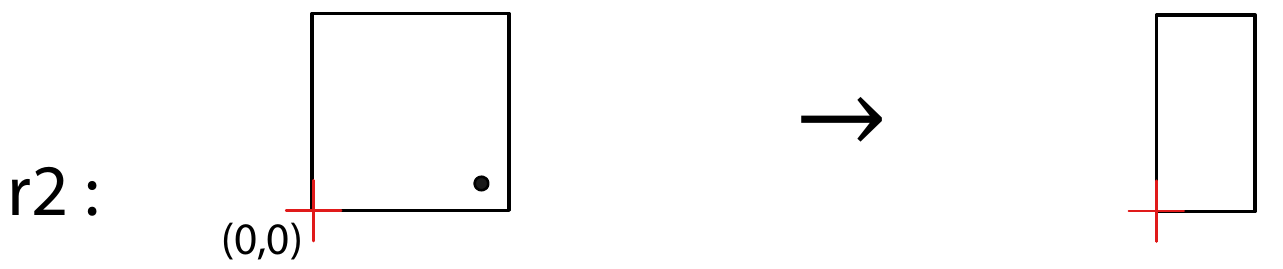}
\label{Figure3}
\end{figure}

The label $\bullet$ is a registration mark that constraints the possible ways in which the shape rules may be applied. The 'red cross' indicates a coordinate system relative to the shape (we often ignore this symbol, but it is always assumed). Before explaining further the underlying mechanisms, we show the language of shapes generated by the two shape rules above.\par

\begin{figure}[h!]
\centering
\includegraphics[scale=0.8]{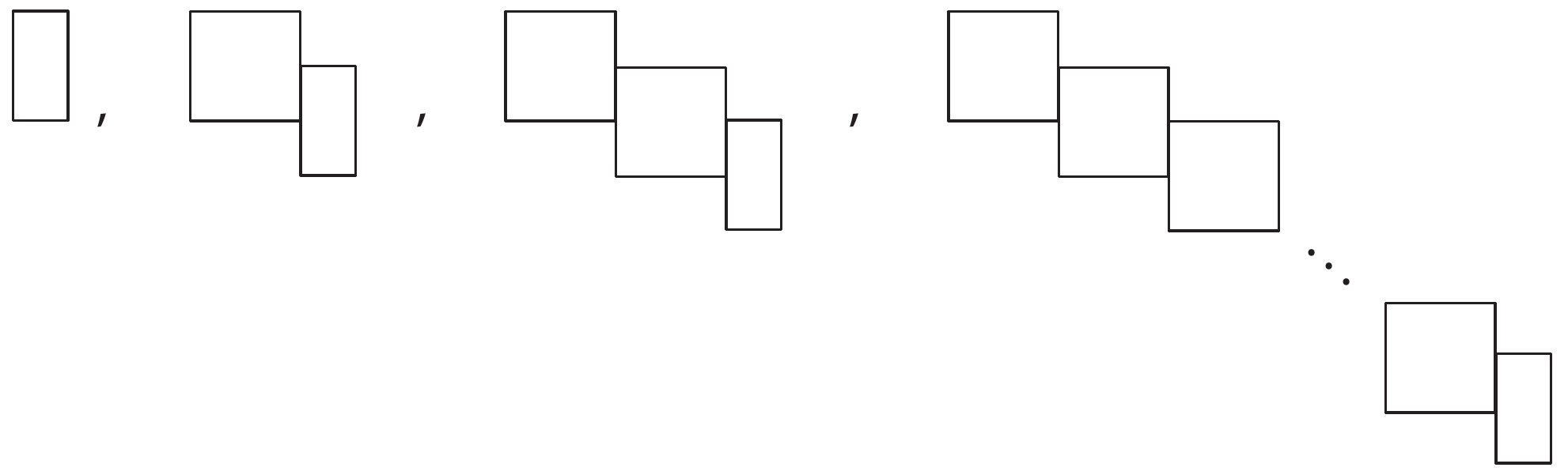}
\label{Figure4}
\end{figure}

\noindent A (shape) rule application works as follows. Let \emph{a} $\rightarrow$ \emph{b} be a rule and \emph{S} an initial shape. For a rule to be applicable to \emph{S}, there must exist a transformation \emph{t} that makes the shape \emph{t}(\emph{a}) a subshape of \emph{S}. The transformations that we deal with here are the Euclidean ones augmented with uniform scale. If a rule is applicable, then the new shape \emph{S'} is computed as follows: (\emph{S} - \emph{t}(\emph{a})) + \emph{t}(\emph{b}). We show the following step by step derivation (each rule application starts in a different row),\par

\begin{figure}[h!]
\centering
\includegraphics[scale=0.40]{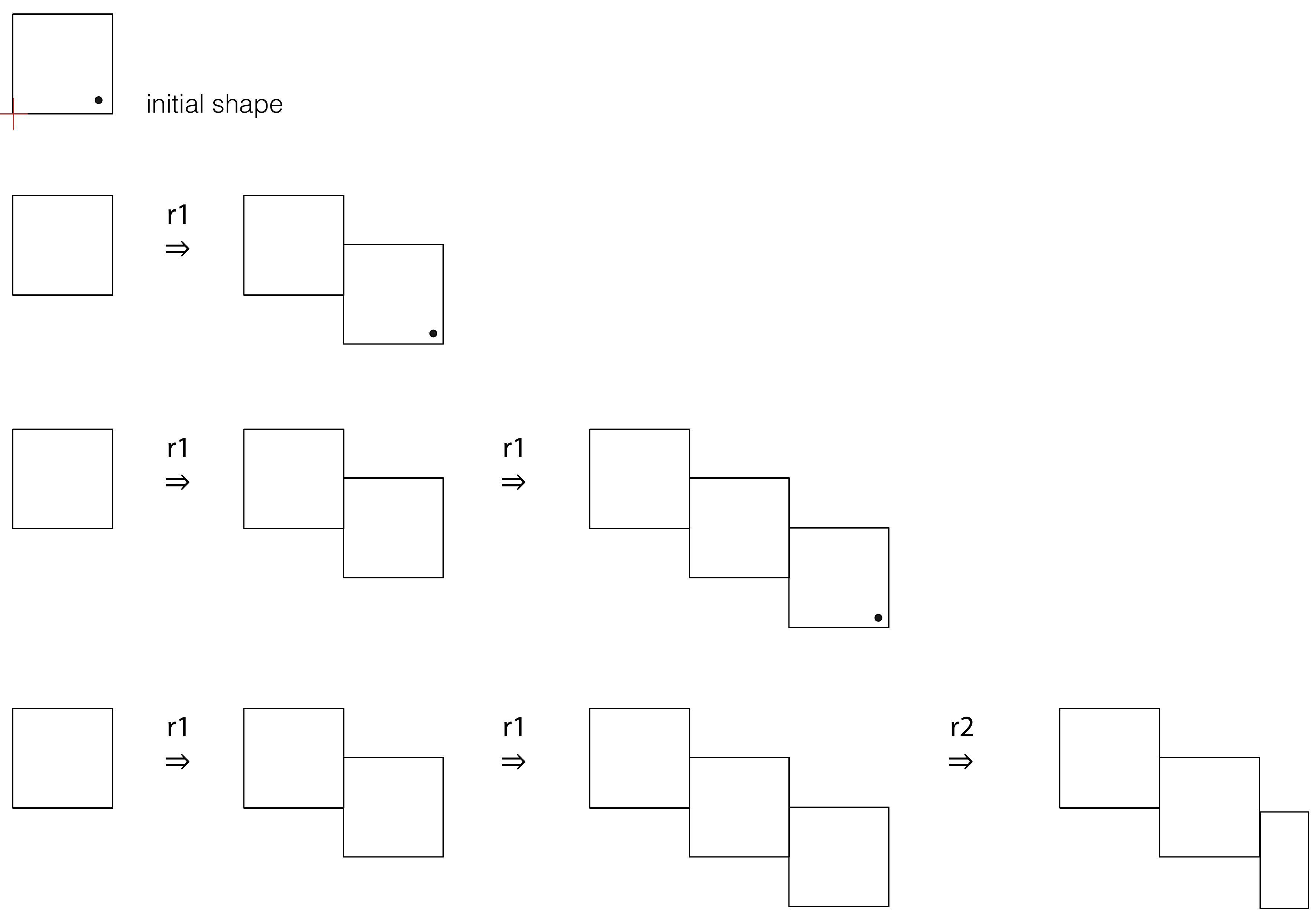}
\label{Figure5}
\end{figure}
    
\noindent The rule applications stop either when there is no other rule applicable or no label $\bullet$. A more elaborate specification of shape rules is the following, which indicates several properties of the left and right hand shapes. 

\begin{figure}[h!]
\centering
\includegraphics[scale=0.6]{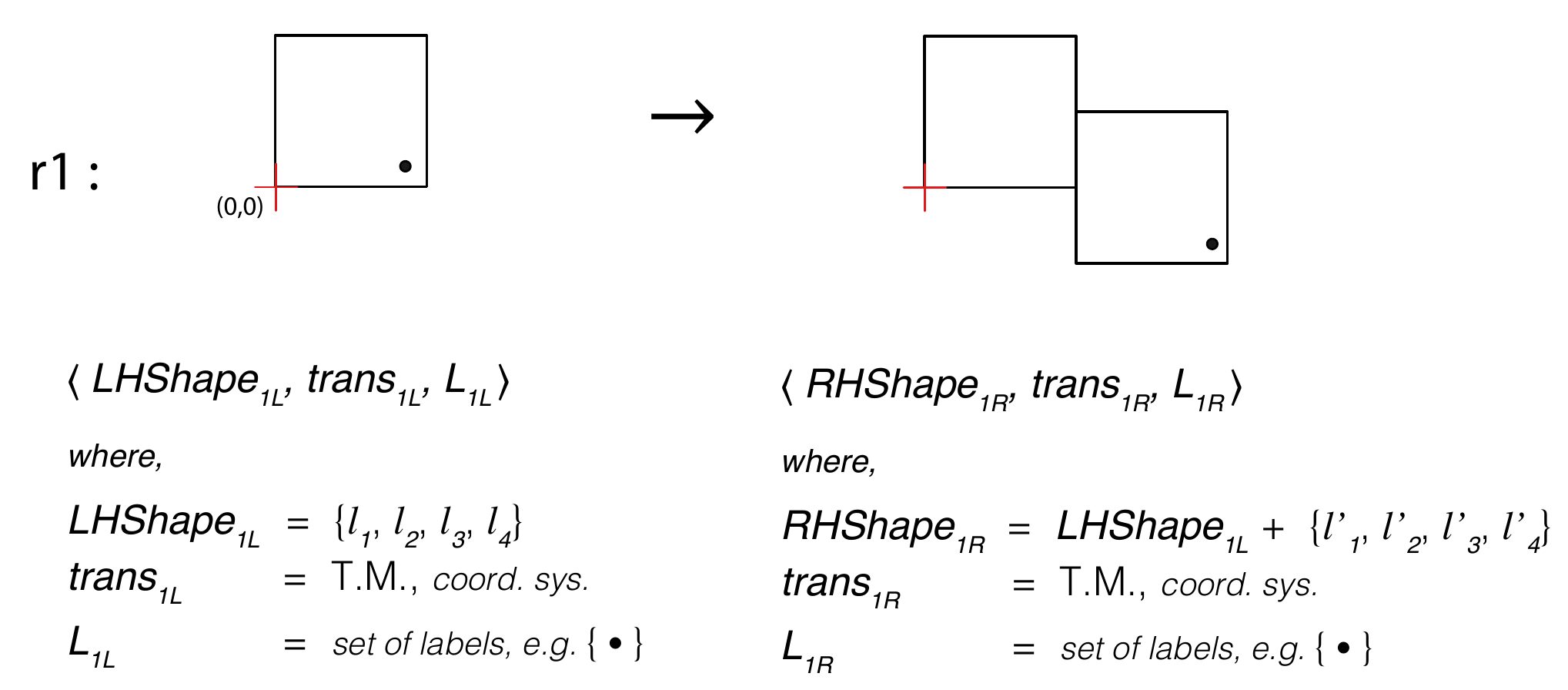}
\label{Figure6}
\end{figure}

More specifically, a labelled shape is a tuple $\langle shape, trans, L \rangle$, where \emph{shape} is a list of linear elements, \emph{trans} is a coordinate system and a transformation matrix describing the translation, rotation, and scale of the shape, and \emph{L} is a list of labels associated with the \emph{shape}; in this case, \emph{L} simply consists of the label $\bullet$. We show this more analytic specification of shape rules in order to illustrate how one may go about implementing them. However, from now on we will only specify shape rules by showing only the left and right hand shapes and ignore the rest of the details.\par 

We give two additional examples of shape rules and their associated languages of shapes. In particular, the following shape rules,

\begin{figure}[h!]
\centering
\includegraphics[scale=0.6]{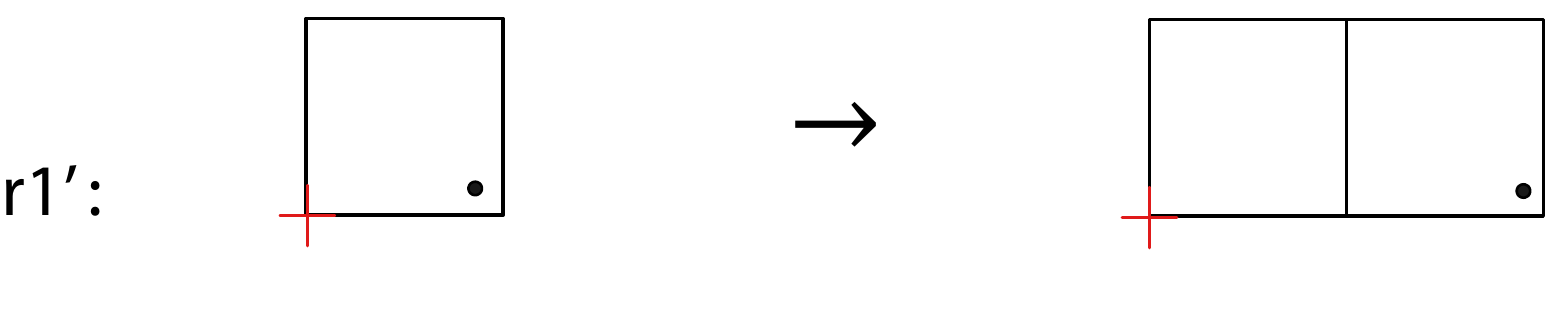}
\includegraphics[scale=0.6]{shape_rule2.pdf}
\label{Figure7}
\end{figure}

\noindent generate the language of shapes,

\begin{figure}[h!]
\centering
\includegraphics[scale=0.8]{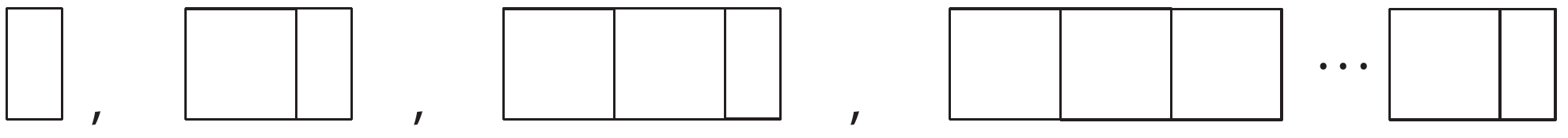}
\label{Figure8}
\end{figure}

\noindent and the next two shape rules,

\begin{figure}[h!]
\centering
\includegraphics[scale=0.6]{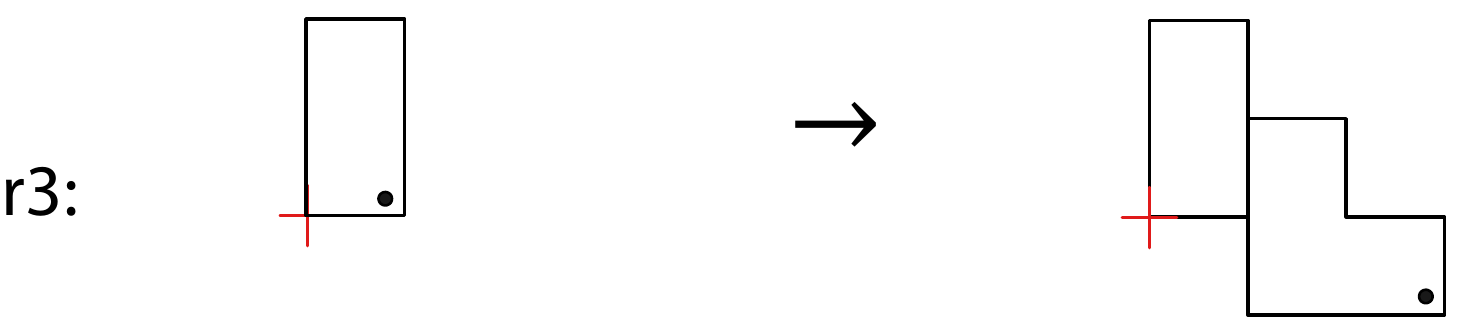}
\includegraphics[scale=0.6]{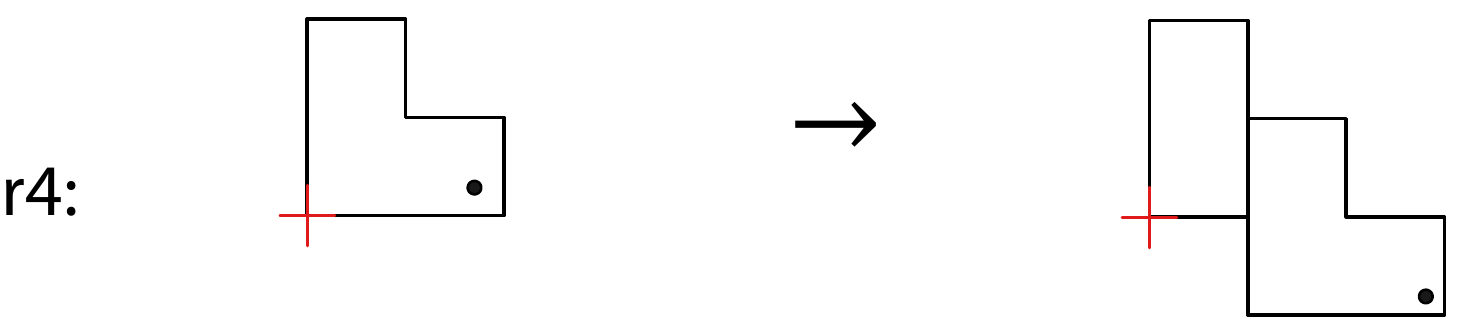}
\label{Figure9}
\end{figure}

\noindent generate the language of shapes on the opposite page.

\begin{figure}[h!]
\centering
\includegraphics[scale=0.8]{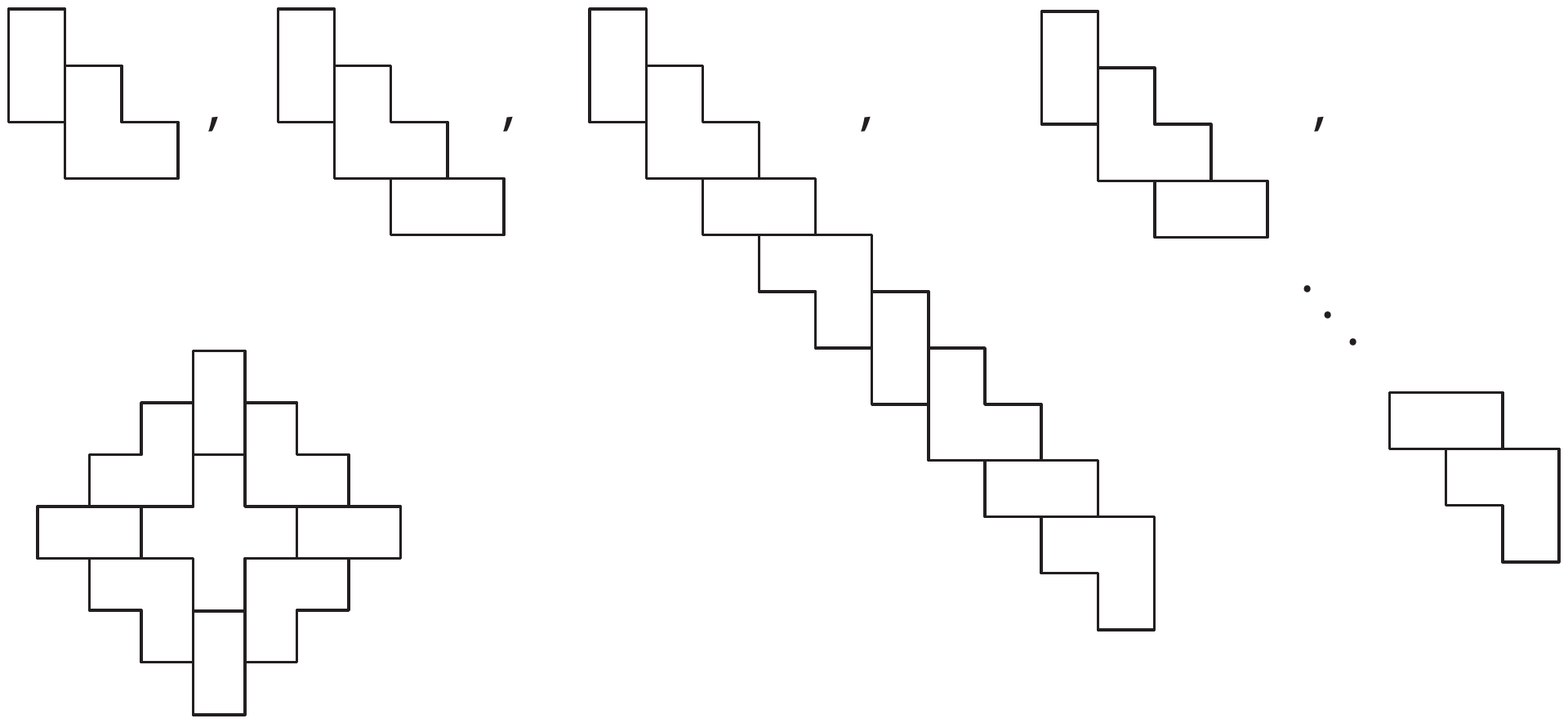}
\label{Figure10}
\end{figure}

The chosen shape rules and their associated languages of shapes are characterized by a certain visual simplicity. We have made this choice to simplify the cases we examine but also because even with these simple arrangements of shapes the very task of describing them in natural language is a hard one. Thus, even with the chosen arrangements, we gain good insights into some of the difficulties involved in connecting perception with natural language.

\subsection{Verbal rules}
\label{sec1}
\noindent In this subsection we discuss the approach we take towards constructing natural language descriptions for arrangements of shapes. Before we do so, however, it is important to emphasize that when it comes to shapes, there exist several approaches that one can take to construct a description for the shapes. The most common description appears in Computer Aided Design where shapes are always represented with certain primitives that have specific geometric characteristics. For example, consider the following arrangement of shapes,

\begin{figure}[h!]
\centering
\includegraphics[scale=0.8]{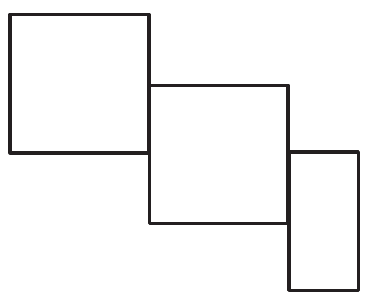}
\label{Figure11}
\end{figure}

\noindent could be described as a set of three geometric primitives, two squares and one rectangle, with specific sizes, i.e. \{\emph{square1}, \emph{square2}, \emph{rectangle1}\}. The individual primitives could be further described with respect to their geometric characteristics, for example in the case of the rectangle \{\emph{rectangle1}, \emph{width} = 1, \emph{length} = 3\}. Another possible way of describing shapes is by providing the sequence of rules that generate them, i.e., a \emph{procedural description}. For example, the following arrangements come with their respective procedural descriptions (rules r3 and r4 are given in subsection 3.1),\par

\begin{figure}[h!]
\centering
\includegraphics[scale=0.6]{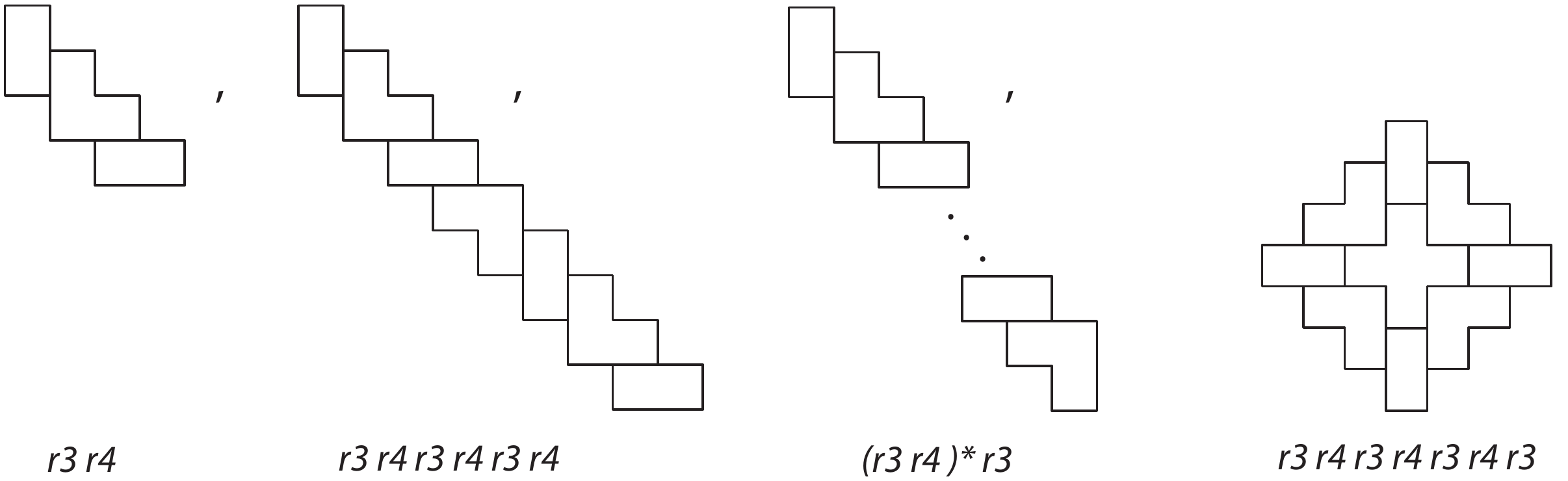}
\label{Figure12}
\end{figure}

\noindent A geometric description and a procedural description are two common ways of describing arrangements of shapes. However, none of the two is a natural language description. Furthermore, in the case of the procedural description, notice that the second arrangement and the fourth arrangement have the same procedural description but their 'appearances' are different. Our goal in this project is to instead construct natural language descriptions, particularly in English, that capture semantically how an arrangement of shapes 'looks'; as if we were to describe the arrangement to a person on the phone.\par

To this end, we propose a system that works with both \emph{shape rules} and, what we call, \emph{verbal rules}. A shape rule shows visually how a shape is transformed to another shape by \emph{replacement}, \emph{addition}, or \emph{subtraction}. For example, rules $r1, r1', r3, r4$ are additive rules; rule $r2$ is a replacing rule (these rules are shown in subsection 3.1). The right hand side of a shape rule always consists of a \emph{spatial relation between two shapes}; it shows pictorially how two shapes are put together. A verbal rule describes in natural language the \emph{action} of the shape rule. Consider the following pair \{\emph{r1}, \emph{v1}\}, where \emph{r1} is a shape rule and \emph{v1} is a verbal rule:\par

\begin{figure}[h!]
\centering
\includegraphics[scale=0.5]{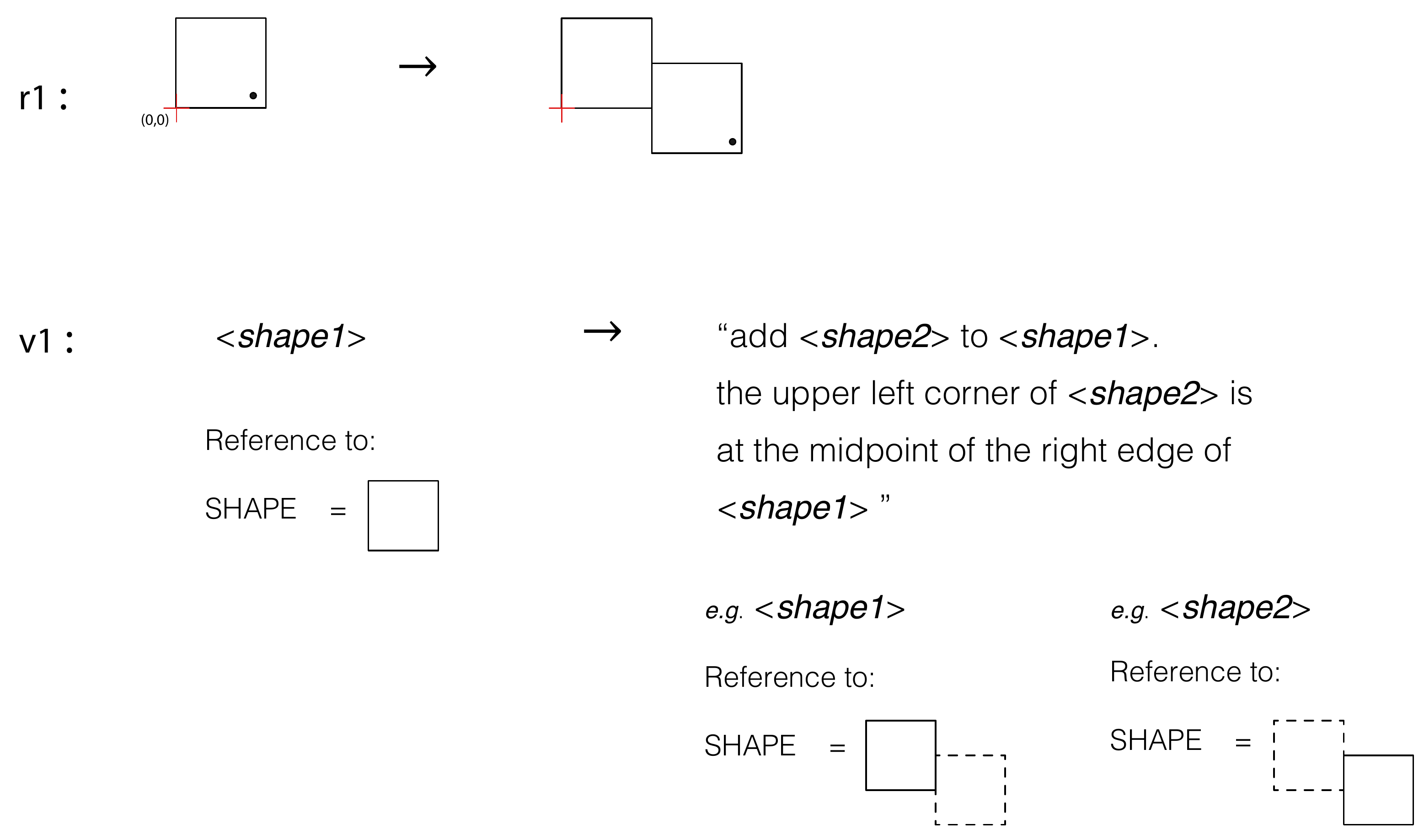}
\label{Figure13}
\end{figure}

\noindent A verbal rule has the general form \emph{v} $\rightarrow$ \emph{w}, where \emph{v} and \emph{w} are strings with references to specific shapes taken from the corresponding shape rule. For example, $<shape1>$ and $<shape2>$ in the verbal rule v1 shown in the above figure are references to the squares in the shape rule r1. The right hand side of a verbal rule always describes \emph{the spatial relation at the right hand side of its associated shape rule}. The two are in correspondence. \par
A shape rule and a verbal rule are applied in parallel. The shape rule generates the arrangement of shapes visually, while the verbal rule describes the action of the shape rule, in effect, describing the generated arrangement of shapes. Notice that in the above example of a verbal rule, the right hand side consists of both a constructive ("add ... to...") and a from-above ("the... is...") description. When a verbal rule is applied to create a natural language description of a shape or an arrangement of shapes, our system semantically interprets the generated description by following the basic spatial semantic concepts we outlined in section 2. We describe the precise underlying mechanisms for how our system parses the natural language description into spatial semantic concepts in section 5. Before we do so, however, we motivate the reader with an example.  

\section{A motivating example}
\label{sec1}

\noindent Consider once again the pair \{\emph{r1}, \emph{v1}\} shown in the previous section. Our system takes as an input the natural language description generated by verbal rule \emph{v1}. The system then assigns spatial roles to words using the spatial semantics concepts outlined in section 2. For the case of the description generated by verbal rule \emph{v1}, following is the semantic interpretation that our system generates:\par 

\begin{figure}[h!]
\centering
\includegraphics[scale=0.42]{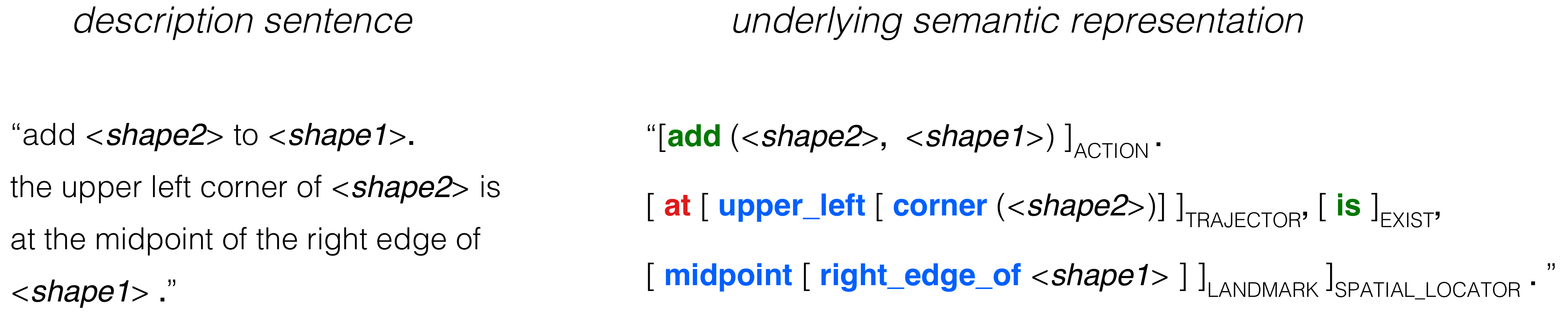}
\label{Figure14}
\end{figure}

\noindent The system understands that the rule is "additive" and that it adds $<shape2>$ to $<shape1>$. The system then understands that \emph{at} is the spatial locator that determines the spatial relation between the two shapes. Further, the system understands specific geometric attributes of $<shape1>$ and $<shape2>$ and assigns the corresponding spatial roles to the correct words. Once the various assignments of spatial roles are complete, the system has a complete semantic representation of the input description sentence.\par
One may wish to apply rules \emph{r1} and \emph{v1} multiple times, to obtain a derivation of arrangements of shapes and their associated verbal descriptions. In a derivation, the system replaces the referenced links for $<shape1>$ and $<shape2>$ with the shapes that correspond to these references, and as a result, the generated descriptions consist of both words and pictures. The two-step derivation shown on the opposite page illustrates this idea. The generated descriptions work essentially like \emph{instructions}; they explain in both a verbal and a pictorial manner how the arrangement of shapes can be constructed, possibly by a human.

\begin{figure}[h!]
\centering
\includegraphics[scale=0.42]{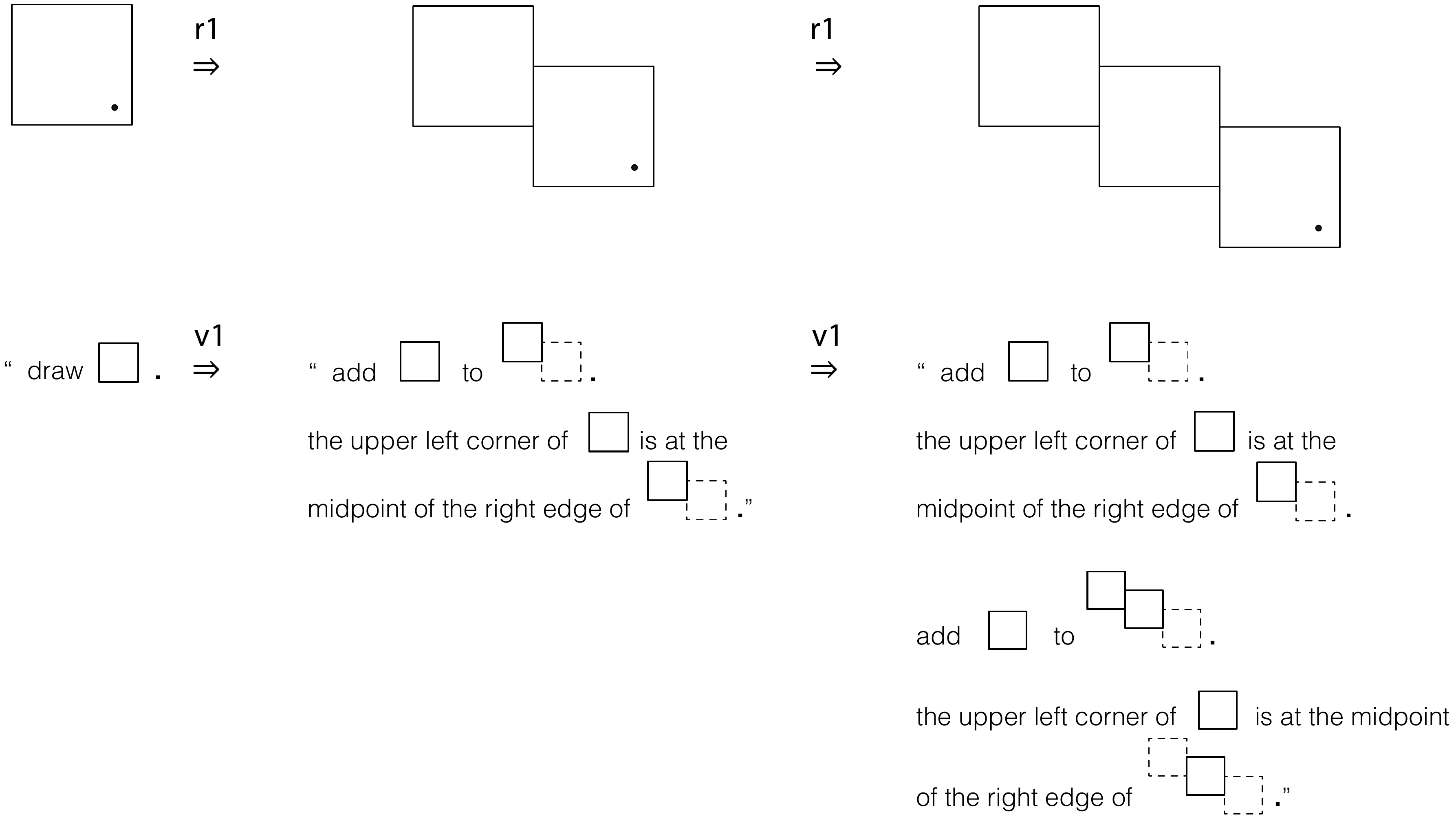}
\label{Figure15}
\end{figure}

\section{Formal spatial semantics representation}
\label{sec1}

\subsection{Semantic interpreter and internal representations}
\label{sec1}

\noindent Our system pairs context-free rules with lambda calculus procedures to generate a spatial semantic interpretation of description sentences. We have a create a set of context-free rules that generate descriptions of arrangements of shapes in natural language and we use lambda calculus compositionality to semantically interpret those sentences using the basic spatial concepts we described in section 2. The complete list of rule-to-rule pairs (i.e. a context-free rule and its associated lambda calculus form) is given in the Appendix. In this section, we explain the principal spatial semantic representations produced by our system using simple description sentences.\par

The system parses an input sentence using context-free rules and generates a parse tree; this parse tree is then passed to a semantic interpreter. The semantic interpreter is equipped with specialized lambda-calculus forms that we developed in order to specifically handle spatial semantics concepts. The semantic interpreter traverses the parse tree in the lambda-calculus fashion and outputs, what we call, a \emph{spatial relation structure} corresponding to a semantic interpretation of the \emph{spatial relations} described in the sentence. The system may also output an \emph{action structure}; these are sentences that start with verbs which we discuss in more detail in the following pages. Both the spatial relation structure and the action structure follow certain templates (see below) with attribute-value pairs. They yield a representation of the meaning conveyed in a given sentence regarding the spatial roles of different words and how these spatial roles are related with one another. We shall give a short example to illustrate how our system generates a spatial semantic structure. Consider the following from-above description sentence and its corresponding spatial semantic structure generated by our system, \par

\vspace{0.2in}

\noindent \emph{"shape1 is at shape2"} \par
\begin{verbatim}
SPATIAL_RELATION['at', "SHAPE['shape1']", 'ttp-nttp',     
                "SHAPE['shape2']", "ACTION['is', 'present']"]
\end{verbatim}

\vspace{0.2in}

\noindent The parse tree of this sentence is: 
\begin{verbatim}
(Start[]
  (S[]
    (NP[-pro, -wh] (SHAPE[] shape1))
    (VP[]
      (EXIST_VERB[] is)
      (PP[] (SP[] at) (NP[-pro, -wh] (SHAPE[] shape2))))))
\end{verbatim}

\vspace{0.1in}

\noindent The internal representation of the above spatial relation structure is based on the following template:

\begin{verbatim}
C("SPATIAL_RELATION", relation=word, region="ttp-nttp", 
            action=action, trajector=trajector, landmark=landmark)
\end{verbatim}

\noindent In essence, the spatial relation structure $"SPATIAL\_RELATION"$ plays the role of an "event" structure (as it is found in the standard lambda calculus approach to semantics). The element that gives the principal meaning in a spatial relation structure is the preposition "at" (and respectively "in" and "on"). Spatial prepositions have the most complex lambda calculus forms and they apply their meaning to the lambda forms of other elements in a sentence. The spatial roles in the above structure are the \emph{trajector}, the \emph{landmark}, the spatial locator \emph{at}, and the topological description \emph{region} based on the regional calculus interpretation of the spatial preposition \emph{at}. Note that the existential verb \emph{is}, in the above sentence, is not a spatial semantics concept. It is only a verb that denotes the existence of shape1 at shape2 (a sample trace of our system assigning spatial semantics role to the from-above description \emph{“shape1 is at shape2”} can be found in the Appendix).\par

Similar to the spatial relation structure, we have an action structure $ACTION$. This applies to constructive sentences of the following form,\par

\vspace{0.2in}

\noindent \emph{add shape1 to shape2}
\begin{verbatim}
ACTION['add', "SHAPE['shape1']", "SHAPE['shape2']"]
\end{verbatim}

\vspace{0.2in}

\noindent In this case, the meaning of the sentence is determined by the action verb "add" which applies its meaning to the other elements, namely, the trajector "shape1" and the landmark "shape2". 
\noindent Now consider the following more complex description sentence:

\vspace{0.1in}

\noindent \emph{“The upper left corner of shape2 is at the midpoint of the right edge of shape1.”} \par

\begin{verbatim}
SPATIAL_RELATION['midpoint', "ACTION['is', 'present']", 
 'ATTRIBUTE["ATTRIBUTE['edge']", "DIRECTION['right']", "SHAPE['shape1']"]', 
        'ttp-nttp', 'at', 'DIRECTION["ATTRIBUTE['corner']", 'upper', 
        "DIRECTION['left']", "SHAPE['shape2']"]']    
\end{verbatim}

\noindent This sentence contains some new spatial roles. Apart from the \emph{trajector}, \emph{landmark}, \emph{region}, and \emph{spatial locator}, there is the spatial concept of an \emph{attribute} of a shape and the spatial concept of a \emph{direction} specific to the \emph{attribute}. Here, again, the existential verb \emph{is} is not a spatial semantics concept. However, the two new spatial concepts, namely \emph{attribute} and \emph{direction}, denote specific geometric properties of the two shapes, i.e. the \emph{trajector} and the \emph{landmark}. \emph{attribute}, as a spatial semantics concept, is a property of a shape, which in this case is either an \emph{edge}, a \emph{corner} of a shape or a \emph{midpoint} of an edge of a shape. \emph{direction} stores the information about the directionality of the \emph{attributes} of the \emph{shapes}. For instance, "the right edge of shape1," may be written in a functional form as \emph{right}(\emph{edge}(\emph{shape1})). Geometric attributes are made with a semantic category with the keyword "ATTRIBUTE" and directions with a semantic category with the keyword "DIRECTION".

A from-above description style can be converted to a constructive description style. For the previous sentence, the constructive description is as follows,\par

\vspace{0.1in}
\noindent \emph{“Draw the upper left corner of shape2 at the midpoint of the right edge of shape1”} \par
\vspace{0.1in}

\noindent In general, our system accepts the following verbs that denote constructive descriptions: \emph{add}, \emph{draw}, \emph{subtract}, and \emph{replace}; these actions correspond, in essence, to shape rules and some examples are given in the next section. These action verbs are put at the start of every new sentence only. The words "shape1" and "shape2" are interpreted as lexical items that follow a noun phrase NP, i.e. $NP \rightarrow SHAPE$ and $SHAPE \rightarrow shape1 \; |\; shape2$ with a lambda procedure corresponding to identity, and a semantic category with the keyword "SHAPE" whose value is determined with whatever shape happens to appear. The spatial prepositions \emph{in} and \emph{on} have the same spatial semantic structure as the preposition \emph{at}. Note that we have omitted to include a special attribute-value pair for the spatial semantic concept \emph{frame of reference} because in the examples we are interested in, shapes are always related with other shapes with respect to a \emph{relative} frame of reference.

\subsection{Description sentences and interpretation: examples}
\label{sec1}

\noindent The system, while limited in many respects, is able to semantically interpret a relatively large spectrum of spatial arrangements of shapes. We will now give a series of examples illustrating how descriptions of different kinds of arrangements of shapes are interpreted by our system. This list is not exhaustive and is only meant to be indicative of the capabilities of the system. \par
In all examples, we show graphically a spatial relation between two shapes (that could be used as the right hand side of a shape rule) and an equivalent verbal description of the arrangement in natural language, either constructive, or from-above or combination of the two styles, along with its semantic interpretation. The existential verb "be" ["is"] is used for all the from-above descriptions and 'action' verbs are used to structure the constructive descriptions. The actions of the constructive descriptions can be additions, subtractions or replacements. It is worth mentioning that the verbal descriptions are made of 24 lexical terms. These are: ["right", "left", "top", "bottom", "upper", "lower" "edge", "corner", "midpoint", "shape1", "shape2", "is", "draw", "add", "subtract", "replace", "at", "on", "in", "to", "from", "with", "of", "the" ].\par

\vspace{0.25in}
\noindent Example 1.\par
\vspace{0.1in}
\noindent \emph{add $<$shape2$>$ to $<$shape1$>$.}\par
\noindent \emph{$<$shape2$>$ is in $<$shape1$>$.}\par

\begin{figure}[h!]
\raggedright
\includegraphics[scale=0.6]{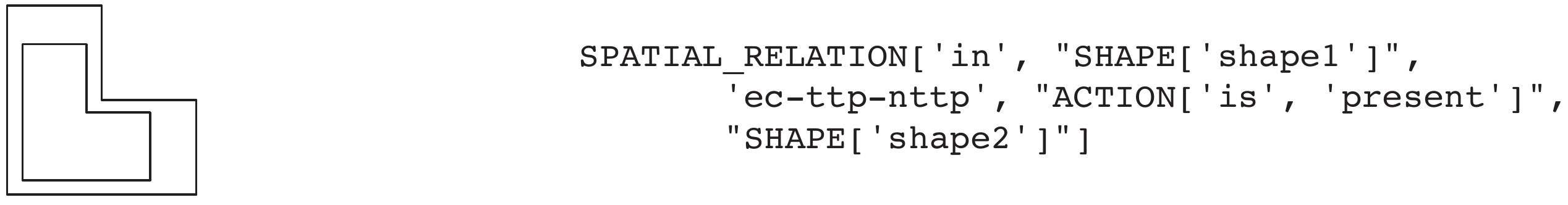}
\label{Figure16}
\end{figure}

\noindent where $<$shape1$>$ and $<$shape2$>$ are the following shapes:\par

\begin{figure}[h!]
\raggedright
\includegraphics[scale=0.6]{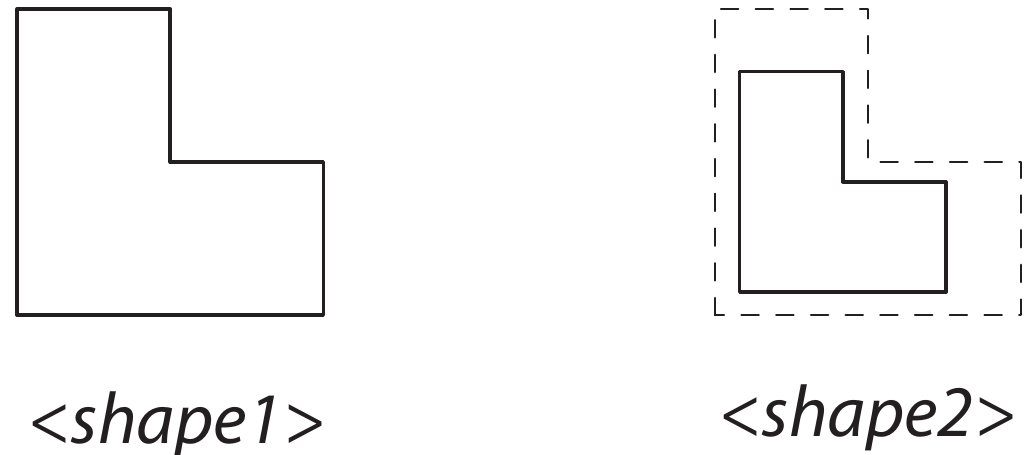}
\label{Figure17}
\end{figure}

\vspace{0.2in}
\noindent Example 2.\par
\vspace{0.1in}
\noindent \emph{add $<$shape2$>$ to $<$shape1$>$.}\par
\noindent \emph{$<$shape2$>$ is on $<$shape1$>$.}\par
\noindent \emph{the bottom edge of $<$shape2$>$ is at the midpoint of the top edge of $<$shape1$>$.}\par

\begin{figure}[h!]
\raggedright
\includegraphics[scale=0.6]{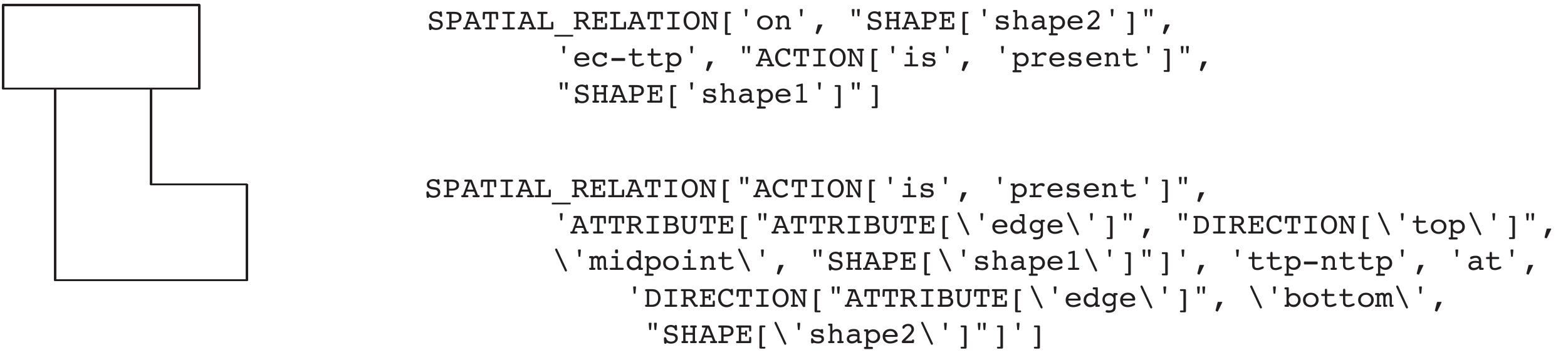}
\label{Figure18}
\end{figure}

\noindent where $<$shape1$>$ and $<$shape2$>$ are the following shapes:\par

\begin{figure}[h!]
\raggedright
\includegraphics[scale=0.6]{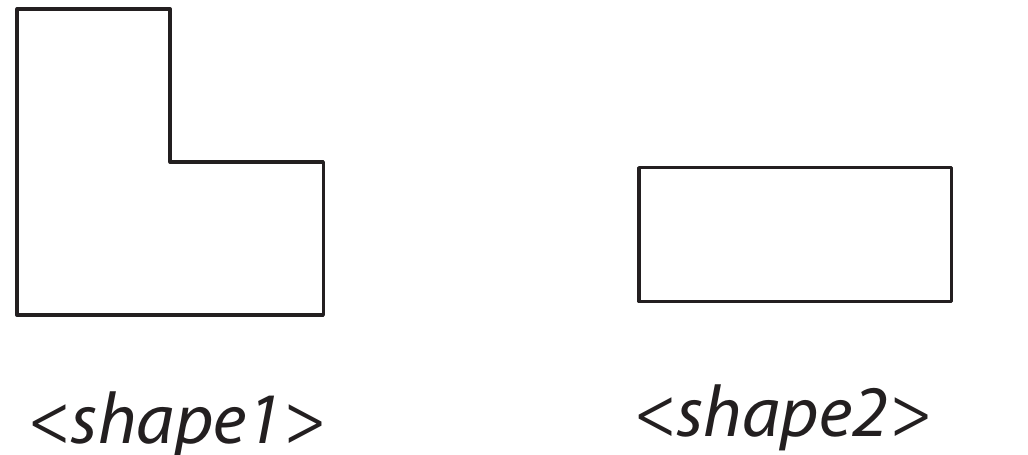}
\label{Figure19}
\end{figure}

\noindent Example 3.\par
\vspace{0.1in}
\noindent \emph{replace $<$shape1$>$ with $<$shape2$>$.}

\begin{figure}[h!]
\raggedright
\includegraphics[scale=0.6]{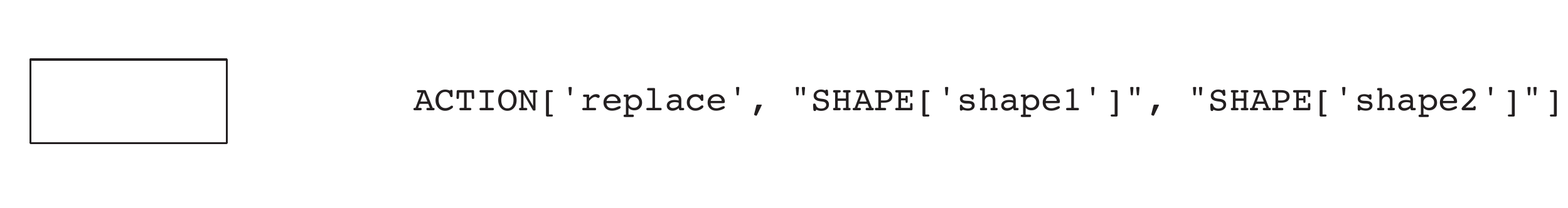}
\label{Figure20}
\end{figure}

\noindent where $<$shape1$>$ and $<$shape2$>$ are the same as in example 2.

\vspace{0.5in}

\noindent Example 4.\par
\vspace{0.1in}
\noindent One may also pursue consecutive arithmetic operations between shapes using verbal descriptions.\par
\vspace{0.1in}
\noindent \emph{add $<$shape1$>$ to $<$shape2$>$.}\par
\noindent \emph{the midpoint of the right edge of $<$shape1$>$ is at the midpoint of the left edge of $<$shape2$>$.}\par

\begin{figure}[h!]
\raggedright
\includegraphics[scale=0.55]{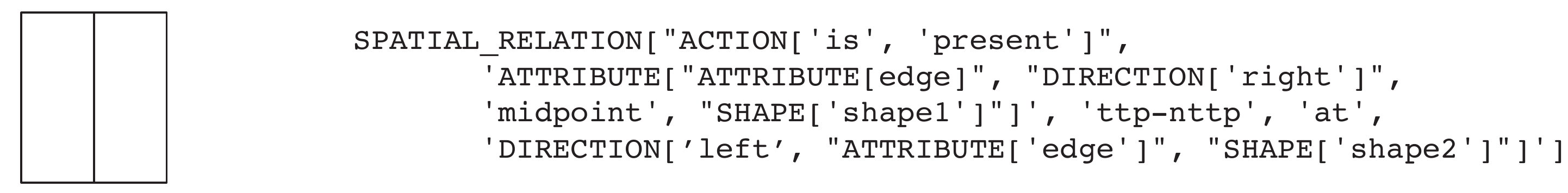}
\label{Figure21}
\end{figure}

\noindent where $<$shape1$>$ and $<$shape2$>$ are the following shapes:\par

\begin{figure}[h!]
\raggedright
\includegraphics[scale=0.5]{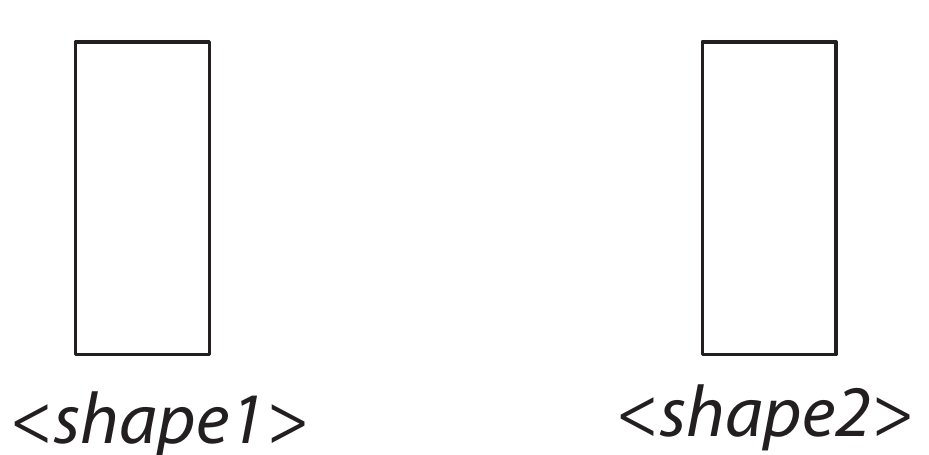}
\label{Figure22}
\end{figure}

\noindent \emph{subtract $<$shape2$>$ from $<$shape1$>$.}\par

\begin{figure}[h!]
\raggedright
\includegraphics[scale=0.5]{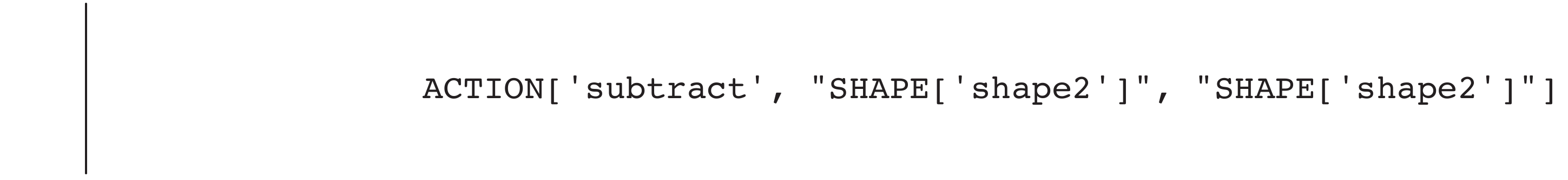}
\label{Figure23}
\end{figure}

\noindent where $<$shape1$>$ and $<$shape2$>$ are the following shapes:\par

\begin{figure}[h!]
\raggedright
\includegraphics[scale=0.5]{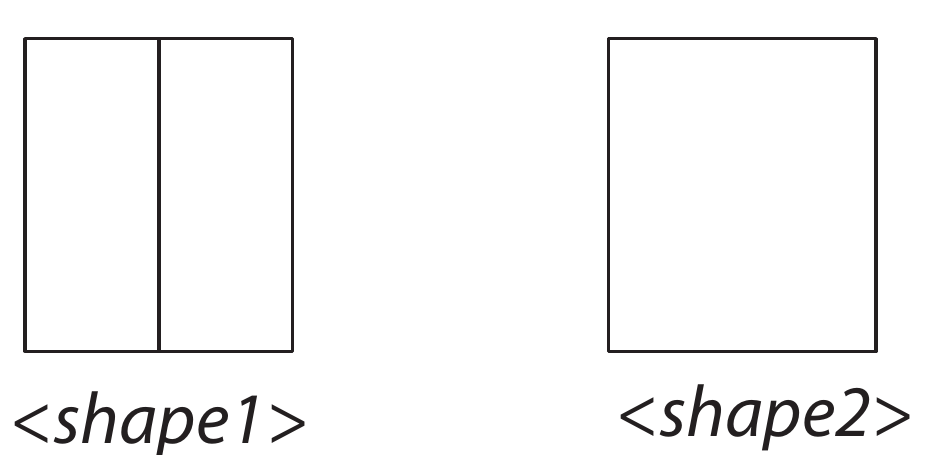}
\label{Figure23}
\end{figure}

\vspace{0.1in}
\noindent Example 5.\par
\vspace{0.1in}
\noindent \emph{the midpoint of the top edge of $<$shape2$>$ is at the midpoint of the right edge of $<$shape1$>$}\par

\begin{figure}[h!]
\raggedright
\includegraphics[scale=0.46]{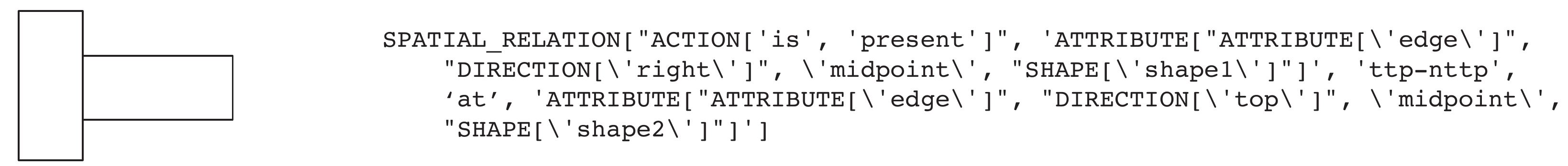}
\label{Figure24}
\end{figure}

\noindent where $<$shape1$>$ and $<$shape2$>$ are the same as in example 4 (first part).
\vspace{0.2in}

\section{Discussion}
\label{sec1}

\noindent Our system successfully performs semantic interpretation of a variety of spatial relations between shapes following the spatial semantics concepts we outlined in section 2. Beyond the current capabilities of the system, there are a number of interesting questions and problems arising, which we briefly discuss in this section.\par

\vspace{0.1in}
\noindent 1. The words $<$shape1$>$ and $<$shape2$>$ match to individual shapes (as is the case with all examples in the previous subsection). However, $<$shape1$>$ and $<$shape2$>$ may also match with any arrangement of shapes that becomes useful for interpreting the action of a shape rule in the course of a computation that proceeds with shape rules and verbal rules jointly applied. \par

In particular, during a derivation with shape rules, at each step, the shape that is being "matched" in a rule application matches under some transformation of the left shape of the shape rule that is being applied at that step. If \emph{a} is this shape, \emph{t}(\emph{a}) is its transformed version (\emph{t} can be an identity, a translation, a rotation, uniform scale or a composition of them). \emph{t}(\emph{a}) matches some part of the \emph{current} arrangement of shapes. If \emph{S} is the current arrangement of shapes, when the rule is applied, \emph{t}(\emph{a}) is subtracted from \emph{S} and a new shape is added, namely, the right hand side of the shape rule \emph{t}(\emph{b}). Thus, at each step, we can always obtain \emph{two} shapes for semantic interpretation as a consequence of a rule application: \emph{t}(\emph{b}) and \emph{S} - \emph{t}(\emph{b}). The semantic interpreter binds $<$shape1$>$ to \emph{S} - \emph{t}(\emph{b}) and $<$shape2$>$ to \emph{t}(\emph{b}) (or the converse) but its "oblivious" as to how exactly these shapes 'look' spatially. To illustrate this concept, imagine two "snapshot" steps within a larger computation (opposite page), where in the first snapshot, at step 1, the word $<$shape1$>$ matches a single shape, whereas in the second snapshot, at some later step \emph{n}, the word $<$shape1$>$ matches the arrangement of shapes that has been created up until step \emph{n} (i.e., shapes created in the previous \emph{n}-1 steps).\par

\begin{figure}[h!]
\centering
\includegraphics[scale=0.45]{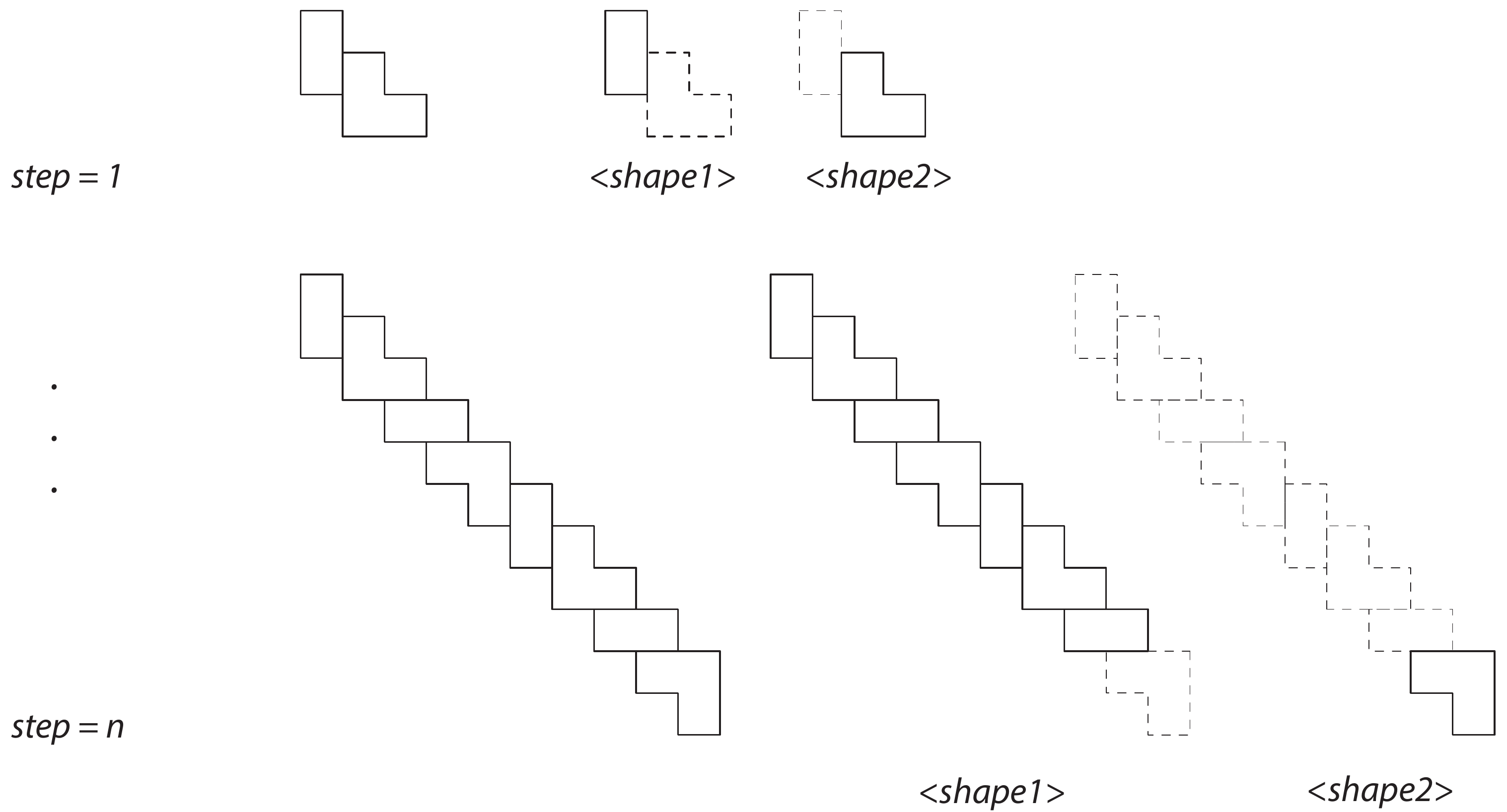}
\label{Figure25}
\end{figure}

This observation is useful because it supports the way we have implemented our spatial semantics interpretation where spatial relations always consist of two participant members: a \emph{trajector} and a \emph{landmark}. Thus, by virtue of the way shape rule applications work, the two spatial roles \emph{trajector} and \emph{landmark} will always correspond to some shape (or arrangement of shapes) pictorially. \par

\vspace{0.25in}

\noindent 2. The fact that at each step of the computation we have two participating shapes \emph{t}(\emph{b}) and \emph{S} - \emph{t}(\emph{b}), raises interesting questions with respect to the possible natural language descriptions that one may pursue. In particular, it would be interesting to explore how a verbal description at some step \emph{n} can be modified with reference to past actions. That is to say, it would be interesting to explore higher-level descriptions that recapitulate or even make use past facts of a computational history to make a more refined (and possibly shorter) verbal description. Consider the arrangement of shapes in the above Figure. One may describe this arrangement not as a monotonically increasing sequence of repeated descriptions of the like, ``shape1 is added to shape2. the lower left corner of shape1 is at....", but a more refined description that backwards references a specific action meant to be repeated multiple times. For example, one possible way to approach this could be to construct descriptions of the sort, ``add shape2 at the midpoint of shape1, \emph{do this} for any newly added shape". The words \emph{do this} backwards reference a specific action described in the first portion of the sentence thus rendering unnecessary its exact repetition. This approach would come closer to the linguistic concept of ``anaphora" but applied to arrangements of shapes.

\vspace{0.25in}
\noindent 3. Apart from constructive and from-above descriptions, humans may describe spatial objects like shapes in many other ways. \emph{Figures of speech}, such as analogies and similarities, and \emph{historical or disciplinary contexts} are two other possible approaches that one may follow to build verbal descriptions of spatial arrangements of shapes. For example, one may describe the following shapes,\par

\begin{figure}[h!]
\centering
\includegraphics[scale=0.25]{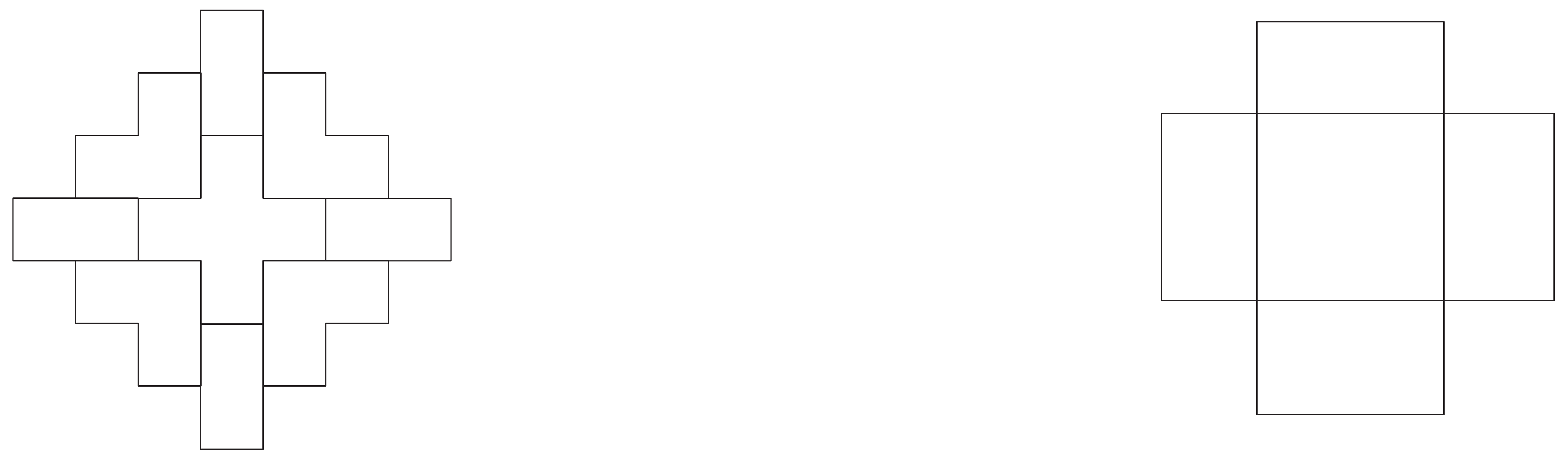}
\label{Figure26}
\end{figure}

\noindent with a verbal description, such as "It looks like a large cross with a smaller cross in the middle" for the first shape (left) and for the second shape (right), "it is a parti used for buildings of classical architecture"; the former incorporates an analogy, the latter incorporates historical and disciplinary knowledge. To represent the semantics of such sentences following lambda calculus compositionality approaches is an interesting future endeavor that would enhance the current system with description styles that are close to what humans would 'talk' about shapes and spatial arrangements of them. 

\vspace{0.25in}
\noindent 4. Another important issue to raise is about the inherent visual ambiguity of shapes. As an example, consider the following shape and its various 'emergent' pieces highlighted with darker outline.

\begin{figure}[h!]
\centering
\includegraphics[scale=0.23]{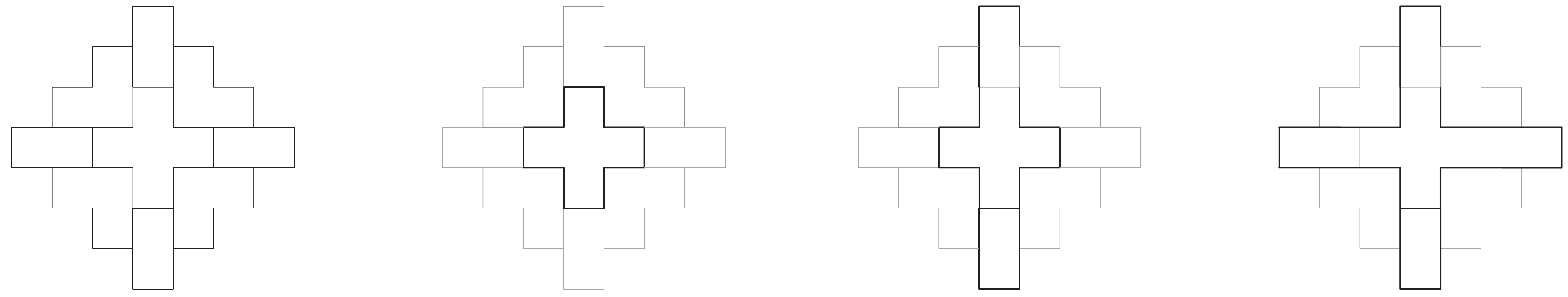}
\label{Figure27}
\end{figure}

\noindent Even if one were to obtain a detailed, constructive description of the arrangement on the far left, the actual generated descriptions would not be able to capture its various emergent parts; those emergent parts are not captured by the underlying representation. These parts, however, are easy to see with a human eye and they may contribute to how one understands the meaning(s) of the arrangement.

\vspace{0.25in}

\noindent 5. One possible extension of our system could be a Graphical User Interface that, having parsed a description sentence into its various spatial roles, it could replace the references to shapes, e.g. $<$shape1$>$ and $<$shape2$>$, with their equivalent graphical depictions. The result of this would be a description that combines natural language with pictures and one motivating example of this was shown earlier in section 4. Moreover, such a graphical interface could help a human disambiguate references to directionality or geometric attribute, e.g. \emph{the right edge} of $<$shape1$>$ or even more specifically \emph{the midpoint of the right edge} of $<$shape1$>$.

\section{Acknowledgements}
\label{sec1}

\noindent The lambda calculus interpreter that we used to build our spatial semantics system on was taken from the course material of the class 6.863J/9.611J, Intro to Natural Language Processing taught by Prof. Robert Berwick at the Massachusetts Institute of Technology in the Spring semester of 2018. We would like to thank Sagar Indurkhya for the useful discussions and helpful feedback.

\vspace{0.4in}
%%%%

\noindent \textbf{Appendix}

\begin{longtabu} to \textwidth { | X[l] | X[l] | }
 \hline
 Context-free Syntactic Rule & Corresponding Semantic Rule \\
 \hline
 Start $\rightarrow$ S  & lambda s: processSentence(s) \\
 \hline
 S $\rightarrow$ NP VP  & lambda np, vp: vp(np) \\
 \hline
 S $\rightarrow$ VACT NPP  & lambda vp, npp: npp(vp) \\
 \hline
 S $\rightarrow$ ADD NP NPP1  & lambda vp, np1, np2: vp(np1, np2) \\
 \hline
 S $\rightarrow$ DRAW NP NPP1  & lambda vp, np1, np2: vp(np1, np2) \\
 \hline
 S $\rightarrow$ SUBTRACT NP NPP1  & lambda vp, np1, np2: vp(np1, np2) \\
 \hline
 S $\rightarrow$ REPLACE NP NPP1  & lambda vp, np1, np2: vp(np1, np2) \\
 \hline
 NP $\rightarrow$ SHAPE  & identity \\
 \hline
 NP $\rightarrow$ DET NP  & lambda det, np: np \\
 \hline
 NPP $\rightarrow$ NP PP  & lambda np, pp: pp(np) \\
 \hline
 NPP1 $\rightarrow$ To NP  & lambda t, np: np \\
 \hline
 NPP1 $\rightarrow$ FROM NP  & lambda t, np: np \\
 \hline
 NPP1 $\rightarrow$ WITH NP  & lambda t, np: np \\
 \hline
 NP $\rightarrow$ PROP OF SHAPE  & lambda pr, of, s: pr(s) \\
 \hline
 PROP $\rightarrow$ DIR ATTR$\_$  & lambda d, a: d(a) \\
 \hline
 PROP $\rightarrow$ COMP DIR$\_$ ATTR$\_$  & lambda c, d, a: c(d, a) \\
 \hline
 PROP $\rightarrow$ ATTR OF DET DIR$\_$ ATTR$\_$  & lambda atr1, o, t, d, atr2: atr1(d, atr2) \\
 \hline
 "DIR", ['right', 'left', 'top', 'bottom']  & lambda word: lambda attr: lambda shape:
                             C("DIRECTION", shape=shape, attribute=attr, direction=word)) \\
 \hline
 "ATTR$\_$", ['edge', 'corner', 'midpoint']  & lambda attribute: 
                             C("ATTRIBUTE", attribute=attribute) \\
 \hline
 "DIR$\_$", ['right', 'left', 'top', 'bottom']  & lambda attribute: 
                             C("DIRECTION", attribute=attribute) \\
 \hline
 "COMP", ['top', 'bottom', 'upper', 'lower']  & lambda word: lambda direct, attr: lambda shape:
                             C("DIRECTION", shape=shape, attribute=attr, direction=direct, comparative=word)) \\
 \hline
 "ATTR", ['edge', 'corner', 'midpoint']  & lambda word: lambda direct, attr: lambda shape:
                             C("ATTRIBUTE", select=word, shape=shape, attribute=attr, direction=direct) \\
 \hline
 VP $\rightarrow$ EXIST$\_$VERB PP  & lambda v, pp: pp(v) \\
 \hline
 VACT $\rightarrow$ ACTION$\_$VERB  & identity \\
 \hline
 "EXIST$\_$VERB", ['is']  & lambda action: C("ACTION", action=action, tense='present') \\
 \hline
 "ACTION$\_$VERB", ['draw', 'add', 'subtract', 'replace']  & lambda action: C("ACTION", action=action, tense='present' \\
 \hline
 ADD $\rightarrow$ 'add'  & lambda word: lambda landmark, trajector:\
		     C("ACTION", action=word, trajector=trajector, landmark=landmark) \\
 \hline
 SUBTRACT $\rightarrow$ 'subtract'  & lambda word: lambda landmark, trajector:\
		     C("ACTION", action=word, trajector=trajector, landmark=landmark) \\
 \hline
 DRAW $\rightarrow$ 'draw'  & lambda word: lambda landmark, trajector:\
		     C("ACTION", action=word, trajector=trajector, landmark=landmark) \\
 \hline
 REPLACE $\rightarrow$ 'replace'  & lambda word: lambda landmark, trajector:\
		     C("ACTION", action=word, trajector=trajector, landmark=landmark) \\
 \hline
 PP $\rightarrow$ IN NP  & lambda p, np: p(np) \\
 \hline
 "SHAPE", ['shape1', 'shape2']  & lambda shape: C("SHAPE", shape=shape) \\
 \hline
 IN $\rightarrow$ 'at'  & lambda word: lambda landmark: lambda action: lambda trajector:\
		        C("SPATIAL$\_$RELATION", relation=word, region="ttp-nttp", action=action, trajector=trajector, landmark=landmark) \\
 \hline
 IN $\rightarrow$ 'on'  & lambda word: lambda landmark: lambda action: lambda trajector:\
		        C("SPATIAL$\_$RELATION", relation=word, region="ec-ttp", action=action, trajector=trajector, landmark=landmark) \\
 \hline
 IN $\rightarrow$ 'in'  & lambda word: lambda landmark: lambda action: lambda trajector:\
		        C("SPATIAL$\_$RELATION", relation=word, region="ec-ttp-nttp", action=action, trajector=trajector, landmark=landmark) \\
 \hline
 To $\rightarrow$ 'to'  & lambda word: lambda: None \\
 \hline
 FROM $\rightarrow$ 'from'  & lambda word: lambda: None \\
 \hline
 WITH $\rightarrow$ 'with'  & lambda word: lambda: None \\
 \hline
 OF $\rightarrow$ 'of'  & lambda word: lambda: None \\
 \hline
 DET $\rightarrow$ 'the'  & lambda word: lambda: None \\
 \hline
 identity  & lambda x: x \\
\hline
\end{longtabu}

\vspace{0.2in}

%% References


\bibliographystyle{elsarticle-num}

\bibliography{sample}

\end{document}